\documentclass[12pt,a4paper]{article}

\usepackage[british]{babel}

\usepackage[a4paper,top=2cm,bottom=2cm,left=2.5cm,right=2.5cm,marginparwidth=1.75cm]{geometry}

\usepackage[style=apa, backend=biber]{biblatex} 
\DeclareLanguageMapping{british}{british-apa} 
\DeclareFieldFormat[article]{volume}{\apanum{#1}} 

\usepackage{amsmath}
\usepackage{graphicx}
\usepackage[colorlinks=true, allcolors=blue]{hyperref}
\usepackage{hyperref}
\usepackage[title]{appendix}
\usepackage{mathrsfs}
\usepackage{amsfonts}
\usepackage{booktabs} 
\usepackage{caption}  
\usepackage{threeparttable} 
\usepackage{algorithm}
\usepackage{algorithmicx}
\usepackage{algpseudocode}
\usepackage{listings}
\usepackage{enumitem}
\usepackage{chngcntr}
\usepackage{booktabs}
\usepackage{lipsum}
\usepackage{subcaption}
\usepackage{authblk}
\usepackage[T1]{fontenc}    
\usepackage{csquotes}       
\usepackage{diagbox}

\usepackage{setspace}
\usepackage{titlesec}
\titleformat{\section} 
  {\normalfont\Large\bfseries}{\thesection.}{1em}{}
  
\usepackage{float}   
\usepackage{caption} 



\title{Calibrating Neural Networks' parameters through Optimal Contraction in a Prediction Problem}

\author[1]{Gonzalo Valdes}
\affil[1]{\small Andres Bello University, Fernandez Concha 700, Santiago, Chile}
\affil[*]{Corresponding author: \texttt{gonzalo.valdes.e@unab.cl}}

\date{}  

\begin{document}
\maketitle

\begin{abstract}
This study introduces a novel approach to ensure the existence and uniqueness of optimal parameters in neural networks. 

The paper details how a recurrent neural networks (RNN) can be transformed into a contraction in a domain where its parameters are linear. It then demonstrates that a prediction problem modeled through an RNN, with a specific regularization term in the loss function, can have its first-order conditions expressed analytically. This system of equations is reduced to two matrix equations involving Sylvester equations, which can be partially solved. 

We establish that, if certain conditions are met, optimal parameters exist, are unique, and can be found through a straightforward algorithm to any desired precision. Also, as the number of neurons grows the conditions of convergence become easier to fulfill.

Feedforward neural networks (FNNs) are also explored by including linear constraints on parameters. According to our model, incorporating loops (with fixed or variable weights) will produce loss functions that train easier, because it assures the existence of a region where an iterative method converges.

\end{abstract}


\newpage
\section{Introduction}\label{sec1}

Neural networks, inspired by the structure and function of the human brain, are versatile computational models used in various applications. They have been applied to Image and Speech Recognition (\cite{krizhevsky2012imagenet, hinton2012deep}), Natural Language Processing (NLP) (\cite{vaswani2017attention, devlin2018bert}), Medical Diagnosis (\cite{ronneberger2015u}), Financial Forecasting (\cite{kaastra1996designing, bucci2020realized}), and other tasks (\cite{chen2018rise, zou2019primer}). Overall, neural networks excel in handling complex, nonlinear relationships and large datasets, making them valuable tools across numerous fields and applications.

Neural networks are typically trained through an optimization procedure that involves finding the best possible parameters. Common techniques include error minimization, likelihood maximization, and entropy maximization, among others (\cite{lecun2015deep, goodfellow2014generative}). These techniques are used to define a loss function, which describes the trade-off between achieving better fit and managing bias.

In practice, this goal is usually achieved through Backpropagation, an efficient gradient descent method that utilizes Leibniz's chain rule (\cite{rumelhart1986learning}). Other methods have also been proposed and used to train neural networks, such as Hebbian learning (\cite{hebb1949organization, brunel1996hebbian}), Boltzmann Machines (\cite{ackley1985learning, salakhutdinov2009deep}), Equilibrium propagation (\cite{scellier2017equilibrium, ernoult2020equilibrium}), Evolutionary computation (\cite{angeline1994evolutionary}), NeuroEvolution of Augmenting Topologies (\cite{stanley2002evolving}), and the Forward-forward algorithm (\cite{Hinton2022Forwardforward}).

However, these algorithms have a significant limitation: they do not guarantee finding the global minimum of the loss function (which may have multiple local minima). Moreover, the reasons behind the multiplicity of local minima are not well understood (\cite{li2018visualizing}).

The paper demonstrates that working in a transformed space (the activation domain) and incorporating a novel regularization into the loss function ensures the existence and uniqueness of optimal parameters under general conditions, guarantying that an algorithm converges if those conditions are met.

The structure of the paper is as follows: first, we show neural networks can be transformed into another domain, where a recurrent neural network can be interpreted as a contraction with linear parameters. Next, we show that a prediction problem modeled through a recurrent neural network, with a loss function incorporating a specific regularization term, has first order conditions that can be written analytically. Then, we show the previous system of equations can be reduced to only two matrix equations that involve a particular form of Sylvester equations, and can be partially solved. Then, we show that under certain conditions optimal parameters exist, are unique, and can be found through a simple algorithm to any desired level of precision. Then, we extend this result to include the case where linear constraints on the parameters are added. The model implies feedforward networks are exceptionally difficult to train, because there is no guarantee of a subset where parameters are easy to train. Finally, we conclude with a summary of our findings.

\section{Recurrent neural networks as contractions}\label{sec2}

This section presents the two equations governing the model. The first equation concerns how independent variables $X$ are transformed into another domain, and the second shows how the transformed data $U(X)$ explains a dependent variable $Y$. We borrow our notation from classical regression analysis.

\subsection{Neural Networks in a different domain}\label{subsec1}

Feed-forward Neural Networks (FNN) are composed by neurons. Each neuron $c_{n,i}$ processes inputs from neurons in the previous layer $c_{n-1,j}$ by transforming the data according to weights $w_{n,i,j}$ and its non-linear activation function $f()$. Afterwards, the neuron sends its output to the next layer. The first layer of neurons $c_{1.i}$ receive raw independent variables $x_j$ and the last layer encapsulates the prediction or classification.

Traditionally, this is modeled according to \ref{eq1}.

\begin{equation}
\label{eq1}
    \begin{aligned}
        c_{n,i}=f_{n,i}\left (b_{n,i}+\sum_{j}{w_{n,i,j}\ c_{n-1,j}}\right)\\
        c_{1.i}=f_{1,i}\left(b_{1,i}+\sum_{j}{v_{1,i,j}x_j}\right)
    \end{aligned}
\end{equation}

In contrast, a Recurrent Neural Network (RNN) connects all neurons to each other through its weights $w_{i,j}$, and also all neurons receive raw independent variables $x_j$ that are processed through $v_{i,j}$, as is shown in \ref{eq2}. That is, a RNN is a generalization of a FNN because a FNN is a RNN where some weights $v$'s and $w$'s are set to zero.

\begin{equation}
\label{eq2}
c_i=f_i\left(\sum_{j}{v_{i,j}x_j}+b_i+\sum_{j}{w_{i,j}c_{j}}\right)
\end{equation}

In turn, this can be re-written in a matrix form, as in \ref{eq3}. Here, V and W are the weights in matrix form. Later, this notation will greatly simplify the analysis.

\begin{equation}
\label{eq3}
c=f\left(V^\prime x+b+W^\prime c\right)
\end{equation}

As can be observed, this equation is not linear on weights $V$ and $W$. Yet, by using the transformation $u\equiv\ V^\prime x+b+W^\prime c$ we find \ref{eq4} and are able to study RNNs in a space where the weights are linear. \footnote{Take $u\equiv\ V^\prime x+b+W^\prime c$, so that by definition $u\equiv\ V^\prime x+b+W^\prime f\left(V^\prime x+b+W^\prime c\right)$. We notice the term inside the parenthesis is simply $u$, finding \ref{eq4}. To return, simply calculate $c=f(u)$, again apply the definition $c=f(V^\prime x+b+W^\prime f\left(u\right))$ and notice it can be written as $c=f(V^\prime x+b+W^\prime c)$, \ref{eq3} } We interpret this change of variable as moving from the Neuron Domain to the Activation Domain.

\begin{equation}
\label{eq4}
u=V^\prime x+b+W^\prime f\left(u\right)
\end{equation}

The above can be interpreted as the governing equation of the RNN, where the independent variables $x$ are transformed into $u$, their equivalents in the activation domain. In other words, equation \ref{eq4} is interpreted similarly to how the Fourier transform (or any other transform) would be understood.

The equation in \ref{eq4} refers to one observation, and can be used to find an output once the weights have been calibrated. In contrast, calibration requires stacking many observations together in a single equation (where only the inputs and outputs are known, and the weights must be calculated), and later use some algorithm for the calibration of parameters. 

By transposing equation \ref{eq4} (and abusing the notation a little) we are able to stack all observations in a single matrix equation, in \ref{eq5}. Later, this equation will be used to minimize a loss function using standard matrix techniques.

\begin{equation}
\label{eq5}
U=XV+1b^\prime+F\left(U\right)W
\end{equation}

To assure a continuous and differentiable behavior of $U$ we must impose some requirements. In the following, we assume $\frac{d f_i}{u_i} \in [0,1]$ so that $F()$ is Lipschitz continuous with a Lipschitz constant of 1 or less. We notice that according to Banach's Contraction Mapping Theorem, a solution to \ref{eq5} always exists and is unique when the norm of $W$ is less than one. This fact will help us mold the Statistical Model.

\subsection{Statistical Model}\label{subsec2}

We must now define a loss function to calibrate the RNNs parameters. In the spirit of borrowing from Linear Regression Analysis, we study a loss function given by half the sum of squares of errors ($\frac{1}{2} \varepsilon^\prime \varepsilon$) plus two regularization functions (for $V$ and $W$). Also, as in linear regression, the errors are the difference between a dependent variable $Y$ and a weighted sum of the transformed independent variables $U(X)$. That is, $\varepsilon=Y-U\beta$. In this context the $\beta$ are assumed to be known; $W$ and $V$ will mold the transformed variables $U$, and through that channel affect the regression model. 

Regarding the regularization of $V$, we observe that equation \ref{eq5} does not pose a particular requirement for it to be well behaved. As in Ridge Regression, we will define its regularization as the $L^2$ norm with parameter $\theta_V$, a parable given by $\frac{\theta_V}{2}Tr\left(V^\prime V\right)$. 

In contrast, the regularization of $W$ must induce a well behaved equation in \ref{eq5}, without actually constraining its parameters to particular values. We define $W$'s regularization as a semi-circle with radius $\theta_W$, given by $\frac{1}{2\theta_W}\left(1-\left(1-\theta_W^2Tr\left(W^\prime W\right)\right)^\frac{1}{2}\right)$. 

With these assumptions in mind, our Statistical Model becomes:

\begin{equation}
\label{eq6}
    \begin{split}
        \begin{aligned}
            & \underset{(W,V)}{\text{minimize}}
            & & \frac{1}{2}\ \varepsilon^\prime\varepsilon+\frac{1}{2\theta_W}\left(1-\left(1-\theta_W^2Tr\left(W^\prime W\right)\right)^\frac{1}{2}\right)+\frac{\theta_V}{2}Tr\left(V^\prime V\right) \\
            & \text{subject to}
            & & Y=U\beta+\varepsilon, \\
            &&& U=XV+1b^\prime+F\left(U\right)W
        \end{aligned}
    \end{split}
\end{equation}

Finally, remember that $F(U) =   
\begin{pmatrix}
    f_1(u_{1,1}) & f_2(u_{1,2}) & f_3(u_{1,3}) & \cdots \\
    f_1(u_{2,1}) & f_2(u_{2,2}) & f_3(u_{2,3}) & \cdots \\
    \vdots & \vdots & \vdots & \ddots  
\end{pmatrix}
$. 

Therefore, in this model $\frac{\partial {f_j(U)}_{i,j}}{\partial u_{m,n}} = 0$ if $(i,j) \neq (m,n)$, which allows us to simplify notation and define the reduced matrix of derivatives as $\dot{F}\left(U\right) \equiv  
\begin{pmatrix}
    \frac{d f_1(u_{1,1})}{d u_{1,1}} & \frac{d f_2(u_{1,2})}{d u_{1,2}} & \frac{d f_3(u_{1,3})}{d u_{1,3}} & \cdots \\
    \frac{d f_1(u_{2,1})}{d u_{2,1}} & \frac{d f_2(u_{2,2})}{d u_{2,2}} & \frac{d f_3(u_{2,3})}{d u_{2,3}} & \cdots \\
    \vdots & \vdots & \vdots & \ddots  
\end{pmatrix}$. 

\subsection{First order conditions}\label{subsec3}

Now that our Statistical Model is clearly defined, we study its First Order Conditions and reduce them to a small set of matrix equations.

Our problem can be written down as a Lagrangian:

\begin{equation}
\label{eq7}
    \begin{split}
        \begin{aligned}
            &  \mathcal{L}(W,V,\varepsilon,\ U,\ \lambda,\ \mu)= \\
            & \frac{1}{2}\ \varepsilon^\prime\ \varepsilon+\frac{1}{2 \theta_W}\ (1-(1-\theta_W^2\ Tr(W^\prime\ W))^{1/2})\ )+\frac{\theta_V}{2}\ Tr(V^\prime\ V) \\
            & +Tr\left(\lambda^\prime\left(U\beta+\varepsilon-Y\right)\right)+Tr\left(\mu^\prime\left(U-XV-1b^\prime-F\left(U\right)W\right)\right)
        \end{aligned}
    \end{split}
\end{equation}

The first order conditions of this problem are:

\begin{equation}
\label{eq8}
    \begin{aligned}
        \varepsilon+\lambda=0\\
        \theta_W\left(1-\theta_W^2tr\left(W^\prime W\right)\right)^{-\frac{1}{2}}W-F\left(U\right)^\prime\mu=0\\
        \theta_VV-X^\prime\mu=0\\
        U-XV-1b^\prime-F\left(U\right)W=0\\
        U\beta+\varepsilon-Y=0\\
        \lambda\beta^\prime+\mu-\dot{F}\left(U\right)\circ(\mu W^\prime)=0
    \end{aligned}
\end{equation}

Where <<$\circ$>> denotes the Hadamard product and $\dot{F}\left(U\right)$ is the already defined matrix of derivatives. The first order conditions in \ref{eq8} can be summarized in only two equations

\begin{equation}
\label{eq9}
    \begin{aligned}
        U=\frac{1}{\theta_V}XX^\prime\mu+1b^\prime+ F\left(U\right)W(U,\mu)\\
        \mu=\left(Y-U\beta\right)\beta^\prime+  \dot{F}\left(U\right)\circ(\mu W^\prime(U,\mu))
    \end{aligned}
\end{equation}
Where, according to Lemma 1, $W(U,\mu)=\frac{1}{\theta_W}  \frac{F\left(U\right)^\prime\mu}{\left({1+ \left \| F\left(U\right)^\prime\mu \right \|_F^2}\right)^{\frac{1}{2}}}$ and $\left\| \right\|_F$ is the Frobenius matrix norm.

Lemma 1 can be interpreted as non-Hebbian learning. In Hebbian learning, the connective strength between two neurons $W(U,\mu)$ increases when two neurons fire with high correlation, $F(U)^\prime F(U)$. In contrast, Lemma 1 indicates the connective strength between one neuron and another is proportional to the correlation between the firing rate of the first, $F(U)$, and how much the loss function would improve if we relaxed the firing model of the second (it's shadow price) $\mu$. That is, the strength of connections $W(U,\mu)$ must be proportional to $F(U)^\prime\mu$.

In turn, equation \ref{eq9} indicates that (a) the firing rate $U$ increases when the shadow price $\mu$ is correlated with the data, and (b) the shadow price $\mu$ increases with the unexplained prediction error $Y - U \beta$.

The relationship between artificial firing rate and shadow price is reminiscent of the relationship between biological neurons and its chemical concentrations. For example, in the Hodgkin-Huxley model of neural synaptic activity, the dynamic of action potentials changes with calcium, sodium and potassium concentrations (\cite{hodgkin1952quantitative}), and in turn, new discoveries have shown that calcium, sodium and potassium ion channels can be modified by repetitive synaptic stimulation (\cite{voglis2006role}). It remains to be seen whether the mechanism induced by maximization in our problem has any meaningful resemblance with a biological brain.

We finish this subsection by noticing that, according to Lemma 2, the system of equations in \ref{eq9} can be solved partially:

\begin{equation}
\label{eq10}
    \begin{aligned}
        U_{B1,B2}= B_1 - \left(I+\beta^\prime \beta Q \right)^{-1} Q B_1 \beta \beta^\prime + 
        Q B_2 - \left(I+\beta^\prime \beta Q \right)^{-1} Q^2 B_2 \beta \beta^\prime  \\
        \mu_{B1,B2}= B_2 - \left(I+ \beta^\prime \beta Q \right)^{-1} Q B_2 \beta \beta^\prime -
        B_1 \beta \beta^\prime + \beta^\prime \beta \left(I+\beta^\prime \beta Q \right)^{-1} Q B_1 \beta \beta^\prime
    \end{aligned}
\end{equation}

with 

\begin{equation}
\label{eq11}
    \begin{aligned}
       Q=\frac{1}{\theta_V}XX^\prime \\
       B_1 = 1b^\prime+ F\left(U\right)W(U,\mu) \\
       B_2 = Y \beta^\prime +  \dot{F}\left(U\right)\circ(\mu W^\prime(U,\mu)) 
    \end{aligned}
\end{equation}

The previous equation involves an inverse whose calculation cost can be reduced by using Woodbury's identity: $(I + \frac{1}{\theta_V} \beta^\prime \beta X X^\prime)^{-1}= I - \frac{\beta^\prime \beta}{\theta_V} X (I + \frac{\beta^\prime \beta}{\theta_V} X^\prime X)^{-1} X^\prime$

\subsection{Existence, Uniqueness and an Iterative Method}\label{subsec4}

Now that the reduced first order conditions have been found we turn to the issue of finding the conditions under which optimal parameters exist and are unique.

First, we notice that a solution to \ref{eq9} must exist. We first turn to Lemma 3, which shows $U$ and $\mu$ are bounded in a closed interval. Then, to prove existence, simply refer to Brouwer's fixed point theorem. Alternatively, turn to Weiestrass's Extreme Value Theorem when observing \ref{eq6}.

The following two theorems are an application of Banach's Contraction Mapping Theorem \footnote{Let $(X, d)$ be a complete metric space and let $T : X \xrightarrow{} X$ be a contraction on $X$ such that $d(T(x_1),T(x_2))\le q d(x_1,x_2)$ with $q<1$. Then $T$ has a unique fixed point $x \in  X$ such that $T(x) = x$.}. They study conditions for uniqueness, finding an iterative method that converges to the optimal parameters under certain conditions. The first theorem is concerned with the general conditions for uniqueness. The second focuses on a subset of possible parameters, finding looser conditions.

\textbf{Theorem 1:} \\
Let $\sup \mu$ and $\sup F(U)$ be the supremum of $\mu$ and $F(U)$ and 
\begin{equation}
\label{eq12}
\begin{aligned}
    G_{U,\mu}
   =
   I +
   \begin{pmatrix}
        sup \: F(U) \\
        sup \: \mu
    \end{pmatrix} 
    \begin{pmatrix}
        sup \: F(U) \\
        sup \: \mu
    \end{pmatrix} ^\prime
    \\
    G_{X,\beta} = I + \beta \beta^\prime \bigotimes Q
\end{aligned}
\end{equation}

Then, when $\theta_W>\kappa(G_{X,\beta}) \frac{1 + \beta^\prime \beta+ \left\| Q \right\|}{ \beta^\prime \beta \left\| Q \right\|} \left\| G_{U,\mu} \right\|$ there exists a small enough $\delta$ so that the contraction given by \ref{eq13} converges to a unique fixed point:

\begin{equation}
\label{eq13}
    \begin{aligned}
    T
   \begin{pmatrix}
      U  \\
      \mu 
    \end{pmatrix}
   \ = \;
   (1-\delta)
   \begin{pmatrix}
      U  \\
      \mu
    \end{pmatrix}
   \ + 
   \delta
       \begin{pmatrix}
           U_{B_1,B_2}(U,\mu)
           \\
           \mu_{B_1,B_2}(U,\mu)
        \end{pmatrix}
    \end{aligned}
\end{equation}

Demonstration: \\
Let us use Banach's Contraction Mapping Theorem:

1) As shown in Lemma 3, $U$ and $\mu$ are bounded in a closed interval and since they are matrices of real numbers, this is enough to show completeness. Also, from Lemma 3 we observe $T()$ is a weighted average between two functions that have itself as domain, and thus it also has itself as domain.

2) The following shows that $ \begin{pmatrix} U_{B_1,B_2}(U,\mu) \\ \mu_{B_1,B_2}(U,\mu) \end{pmatrix}$ is Lipschitz continuous. It will be used to find the conditions under which the contraction converges.

According to Lemma 2, equation \ref{eq9} can be written in terms of matrices by using the $vec$ operator and the Kronecker product. As a consequence, we find:

\begin{equation}
\label{eq14}
    \begin{aligned}
    \left\|
       \begin{pmatrix}
           U_{B_1,B_2}(U_1,\mu_1)-U_{B_1,B_2}(U_2,\mu_2)
           \\
           \mu_{B_1,B_2}(U_1,\mu_1)-
           \mu_{B_1,B_2}(U_2,\mu_2)
        \end{pmatrix}
    \right\|
        \le
    \kappa(G_{X,\beta}) \frac{1 + \beta^\prime \beta+ \left\| Q \right\|}{\beta^\prime \beta \left\| Q \right\|}
    \left\|
        \begin{pmatrix}
            B_1(U_1,\mu_1) - B_1(U_2,\mu_2) \\
            B_2(U_1,\mu_1) - B_2(U_2,\mu_2)
        \end{pmatrix}
    \right\|
    \end{aligned}
\end{equation}

Where $\kappa()$ is the condition number.

According to Lemma 4.1, $ \begin{pmatrix} U_{B_1,B_2}(U,\mu) \\ \mu_{B_1,B_2}(U,\mu) \end{pmatrix}$ is Lipschitz continuous with:

\begin{equation}
\label{eq15}
\begin{aligned}
    \left\|
        \begin{pmatrix}
            B_1(U_1,\mu_1) - B_1(U_2,\mu_2) \\
            B_2(U_1,\mu_1) - B_2(U_2,\mu_2)
        \end{pmatrix}
    \right\|
    \le
    \frac{1}{\theta_W}
   \left\| G_{U,\mu} \right\|
   \left\|
   \begin{pmatrix}
         U_1 - U_2  \\
        \mu_1 - \mu_2 
    \end{pmatrix}  
    \right\|
    \end{aligned}
\end{equation}

Composing both inequalities we find that $\begin{pmatrix} U_{B_1,B_2}(U,\mu) \\ \mu_{B_1,B_2}(U,\mu) \end{pmatrix}$ is Lipschitz continuous

\begin{equation}
\label{eq16}
    \begin{aligned}
    \left\|
       \begin{pmatrix}
           U_{B_1,B_2}(U_1,\mu_1)-U_{B_1,B_2}(U_2,\mu_2)
           \\
           \mu_{B_1,B_2}(U_1,\mu_1)-
           \mu_{B_1,B_2}(U_2,\mu_2)
        \end{pmatrix}
    \right\|
        \le 
    \kappa(G_{X,\beta}) \frac{1 + \beta^\prime \beta+ \left\| Q \right\|}{\beta^\prime \beta \left\| Q \right\|}
    \frac{\left\| G_{U,\mu} \right\|}{\theta_W}
   \left\|
   \begin{pmatrix}
         U_1 - U_2  \\
        \mu_1 - \mu_2 
    \end{pmatrix}  
    \right\|
    \end{aligned}
\end{equation}

3) Finally, according to Lemma 5, the previous condition means it will be possible to find $\delta$ small enough so that the contraction in \ref{eq13} converges. That is, we conclude that when
$\theta_W>\kappa(G_{X,\beta}) \frac{1 + \beta^\prime \beta+ \left\| Q \right\|}{ \beta^\prime \beta \left\| Q \right\|} \left\| G_{U,\mu} \right\| $ we can find $\delta$ small enough so that the contraction given by \ref{eq13} converges.

This proves the conditions of Banach's Contraction Mapping Theorem. When the conditions are met the contraction in \ref{eq13} will have exactly one fixed point, and it can be found by simple iteration. That is, when the previous conditions are met the contraction represents an iterative method that always converges to the unique first order conditions of our statistical model. QED.

The conditions in Theorem 1 may require large regularization constants to prevent multiple minima from arising. As the proof of Theorem 1 shows, this is because the weight matrix $W(U,\mu)$ may incorporate positive or negative numbers, so differences between alternative $(U,\mu)$ pairs are not necessarily damped. The following theorem restricts parameters to a subset where differences are softened, finding looser convergence conditions.\\

\textbf{Theorem 2:} \\
Assume an optimal weight matrix $W^*$ exists and an  invertible matrix $\Omega$ exists such that $(\Omega W^* \ge 0$ and $W^* \Omega \ge 0$ (entry wise).
Then, define $S$ as the Subset of pairs $(U,\mu)$ where all the matrices $W(U,\mu)$ are such that $\Omega W(U,\mu) \ge 0$ and $W(U,\mu) \Omega \ge 0$ (entry wise). 

Then, when $\theta_W>\kappa(G_{X,\beta}) \kappa(\Omega) \frac{1 + \beta^\prime \beta+ \left\| Q \right\|}{\beta^\prime \beta \left\| Q \right\|}$ it is possible to find a small enough $\delta$ so that the iterative method given by \ref{eq13} will converge to a unique fixed point in $S$.

Demonstration: \\
Let us use Banach's Contraction Mapping Theorem, again.
First, we notice that $S$ is not void because an optimal weight matrix exists (as Brouwer's Fixed Point Theorem shows) and by assumption it is part of the subset. Also, if $W^*$ is invertible, we can assure a matrix $\Omega$ that meets the conditions exist (simply choose $\Omega=(W^*)^{-1}$). 

Now we notice $S$ is complete, because it is a non-void intersection between complete and convex subsets.

According to Lemma 4.2, we find that in $S$:

\begin{equation}
\label{eq17}
\begin{aligned}
    \left\|
        \begin{pmatrix}
            B_1(U_1,\mu_1) - B_1(U_2,\mu_2) \\
            B_2(U_1,\mu_1) - B_2(U_2,\mu_2)
        \end{pmatrix}
    \right\|
    \le
    \frac{\kappa(\Omega)}{\theta_W}
   \left\|
   \begin{pmatrix}
         U_1 - U_2  \\
        \mu_1 - \mu_2 
    \end{pmatrix}  
    \right\|
    \end{aligned}
\end{equation}

Again, composing inequalities in \ref{eq14} and \ref{eq17}, we find $\begin{pmatrix} U_{B_1,B_2}(U,\mu) \\ \mu_{B_1,B_2}(U,\mu) \end{pmatrix}$ is Lipschitz continuous.

Using Lemma 5 again, we conclude  that when $\theta_W>\kappa(G_{X,\beta}) \kappa(\Omega) \frac{1 + \beta^\prime \beta+ \left\| Q \right\|}{\beta^\prime \beta \left\| Q \right\|}$ we can find a $\delta$ so that the distance between different $(U,\mu)$ pairs diminishes in $S$. 

Also, we notice that since $T()$ is Lipschitz continuous and, since the distance between a contraction and the fixed point always becomes smaller than the distance between the original data and the fixed point, we can find a small enough $\delta$ that guarantees the contraction is also part of $S$. 

This proves Banach's Contraction Mapping Theorem inside $S$, meaning there exists a small $\delta$ where the contraction in \ref{eq13} converges to a unique fixed point. QED.

That is, if $W^*$ is an invertible weight matrix that represents a (possibly not unique) fixed point $(U^*,\mu^*)$ in our problem, we find that in the subset $S$ it is possible to find a small $\delta$ so that the iterative method will converge to exactly one fixed point, $(U^*,\mu^*)$. 

This is a looser condition than the one found in theorem 1, because convergence requires smaller parameters. This is because we have found optimal parameters are surrounded by alternatives that will iteratively converge as long as there is a  matrix $\Omega$ that is invertible.

As a closing remark of this section, we notice that as the number of observations grows the norm of $Q$ diverges, $\left\| Q \right\| \xrightarrow{} \infty$. When this happens the ratio $\frac{1 + \beta^\prime \beta+ \left\| Q \right\|}{\beta^\prime \beta \left\| Q \right\|} \xrightarrow{} \frac{1}{\beta^\prime \beta}$, which means the conditions for convergence become less strict. In particular, if $\beta$s are equal to 1 then $\beta^\prime \beta$ equals the number of neurons and as the number of neurons grows, the looser the conditions of convergence will become.

\section{Constrained RNNs and Feedforward networks}\label{sec3}

Adding constraints allows the modeler to choose an adequate number of parameters. The number of parameters in a complete RNN can turn the estimation problem intractable. Imagine we wanted to train a RNN model with ten thousand neurons to process photos of ten megapixels. Then, we would have to tune matrix $V$ with $10^{11}$ parameters and matrix $W$ with $10^8$; a large number of parameters that may not be supported by the number of datapoints that can lead to excessive adaptation to the training data; overfitting. Constraints in RNNs may help the modeler extract the underlying structure of the data without limiting the number of neurons.

Furthermore, incorporating constraints will allows us to study feedforward neural networks (FNN). An FNN is characterized by the direction of the flow of information between its layers. Its flow is uni-directional, meaning that the information in the model flows from the input nodes, through the hidden nodes and to the output nodes. In contrast to Recurrent Neural Networks (RNN), there aren't cycles or loops which have a bi-directional flow. 

FNNs can be interpreted as constrained RNNs, in which all the loops in hidden layers have weights set to zero, and only input neurons receive the exterior data. Instead of creating a specific model for FNNs, we will solve the general model of a constrained RNN. This section will show that in FNNs it cannot be assured that a Well-Behaved Subset exists, where simple iteration of the contraction will lead towards the optimum parameters.

Two strategies can be used to add linear constraints: a) writing down the constrained variable as part of a hyperplane originating from a normal vector (matrix) and a translation vector (matrix), and b) literally adding constraints to the Lagrangian problem. Both approaches are mathematically equivalent, but one may be more practical than the other in terms of management or complexity of matrix multiplication or matrix inversion.

In the following model we will use both approaches in parallel: matrix $V$ will be characterized according to strategy a) $V = N V_r + V_0$ and matrix $W$ will be constrained according to strategy b) $R \: vec(W) = r$. Please notice the operator $ vec() $ is necessary in order to be able to constrain one component of matrix $W$ without affecting the others.

With these ideas in mind, our Statistical Problem becomes:

\begin{equation}
\label{eq18}
    \begin{split}
        \begin{aligned}
            & \underset{(W,V)}{\text{minimize}}
            & & \frac{1}{2}\ \varepsilon^\prime\varepsilon+\frac{1}{2\theta_W}\left(1-\left(1-\theta_W^2Tr\left(W^\prime W\right)\right)^\frac{1}{2}\right)+\frac{\theta_V}{2}Tr\left(\left( N V_r + V_0 \right)^\prime \left( N V_r + V_0 \right)\right) \\
            & \text{subject to}
            & & Y=U\beta+\varepsilon, \\
            &&& U=XV+1b^\prime+F\left(U\right)W \\
            &&& R \: vec(W) = r
        \end{aligned}
    \end{split}
\end{equation}

Our problem can be written down as a Lagrangian:

\begin{equation}
\label{eq19}
    \begin{split}
        \begin{aligned}
            &  \mathcal{L}(W,V,\varepsilon,\ U,\ \lambda,\ \mu)= \\
            & \frac{1}{2}\ \varepsilon^\prime\ \varepsilon+\frac{1}{2 \theta_W}\ (1-(1-\theta_W^2\ Tr(W^\prime\ W))^{1/2})\ )+\frac{\theta_V}{2}Tr\left(\left( N V_r + V_0 \right)^\prime \left( N V_r + V_0 \right)\right) \\
            & +Tr\left(\lambda^\prime\left(U\beta+\varepsilon-Y\right)\right)+Tr\left(\mu^\prime\left(U-X\left( N V_r + V_0 \right)-1b^\prime-F\left(U\right)W\right)\right)+ Tr \left( \gamma^\prime	(R \: vec(W) - r)\right)
        \end{aligned}
    \end{split}
\end{equation}

The first order conditions of this problem are:

\begin{equation}
\label{eq20}
    \begin{aligned}
        \varepsilon+\lambda=0\\
        \theta_W\left(1-\theta_W^2tr\left(W^\prime W\right)\right)^{-\frac{1}{2}}W-F\left(U\right)^\prime\mu - vec^{-1}\left(R^\prime\gamma\right)=0\\
        \theta_V N^\prime\left( N V_r + V_0\right) -N^\prime X^\prime\mu=0\\
        U-X \left( N V_r + V_0 \right)-1b^\prime-F\left(U\right)W=0\\
        U\beta+\varepsilon-Y=0\\
        \lambda\beta^\prime+\mu-\dot{F}\left(U\right)\circ(\mu W^\prime)=0\\
        R \: vec(W) - r = 0
    \end{aligned}
\end{equation}

As in the previous problem, the first order conditions in \ref{eq20} can be summarized in only two equations

\begin{equation}
\label{eq21}
    \begin{aligned}
        U= \frac{1}{\theta_V} X N \left(N N^\prime \right)^{-1} N^\prime\left( X^\prime\mu - \theta_V V_0\right)+ X V_0+1b^\prime+ F\left(U\right)W(U,\mu)\\
        \mu=\left(Y-U\beta\right)\beta^\prime+  \dot{F}\left(U\right)\circ(\mu W^\prime(U,\mu))
    \end{aligned}
\end{equation}

Where, according to Lemma 6, $W(U,\mu)=\frac{1}{\theta_W} \left( \frac{1-\theta_W^2 \left\| W_2 \right\|_F^2 }{1+ \left\| W_1(U,\mu) \right\|_F^2}\right) W_1(U,\mu)+W_2$ with 

\begin{equation}
\label{eq22}
    \begin{aligned}
        W_1(U,\mu)=vec^{-1}\left(\left[I-R^\prime\left(RR^\prime\right)^{-1} R\right]vec\left(F\left(U\right)^\prime\mu\right)\right)\\
        W_2=\ vec^{-1}\left(R^\prime\left(RR^\prime\right)^{-1}r\right)
    \end{aligned}
\end{equation}

It can be easily observed that if $\left\|W_2 \right\|_F < \frac{1}{\theta_W}$ then $\left\|W \right\| < \frac{1}{\theta_W}$

Again, this equation can be partially solved using Lemma 2:

\begin{equation}
\label{eq23}
    \begin{aligned}
        U_{\hat{B}1,\hat{B}2}= \hat{B}_1 - \left(I+\beta^\prime \beta \hat{Q} \right)^{-1} \hat{Q} \hat{B}_1 \beta \beta^\prime + 
        \hat{Q} \hat{B}_2 - \left(I+\beta^\prime \beta \hat{Q} \right)^{-1} \hat{Q} ^2 \hat{B}_2 \beta \beta^\prime  \\
        \mu_{\hat{B}1,\hat{B}2}= \hat{B}_2 - \left(I+ \beta^\prime \beta \hat{Q} \right)^{-1} \hat{Q} \hat{B}_2 \beta \beta^\prime -
        \hat{B}_1 \beta \beta^\prime + \beta^\prime \beta \left(I+\beta^\prime \beta \hat{Q} \right)^{-1} \hat{Q} \hat{B}_1 \beta \beta^\prime
    \end{aligned}
\end{equation}

with 

\begin{equation}
\label{eq24}
    \begin{aligned}
      \hat{Q}=\frac{1}{\theta_V} X N \left(N N^\prime \right)^{-1} N^\prime\ X^\prime \\
      \hat{B}_1 = X\left(I - N \left(N N^\prime \right)^{-1} N^\prime \right) V_0  + 1b^\prime+ F\left(U\right)W(U,\mu) \\
      \hat{B}_2 = Y \beta^\prime +  \dot{F}\left(U\right)\circ(\mu W^\prime(U,\mu)) 
    \end{aligned}
\end{equation}

Now, we notice theorems 1 and 2 in the previous section are still valid with small modifications. Therefore there is a small $\delta$ such that the contraction given by \ref{eq25} converges to a unique fixed point

\begin{equation}
\label{eq25}
    \begin{aligned}
    T
   \begin{pmatrix}
      U  \\
      \mu 
    \end{pmatrix}
   \ = \;
   (1-\delta)
   \begin{pmatrix}
      U  \\
      \mu
    \end{pmatrix}
   \ + 
   \delta
       \begin{pmatrix}
           U_{\hat{B}_1,\hat{B}_2}(U,\mu)
           \\
           \mu_{\hat{B}_1,\hat{B}_2}(U,\mu)
        \end{pmatrix}
    \end{aligned}
\end{equation}

\begin{enumerate}[label=\alph*)]
    \item in the set of all possible $(U,\mu)$ pairs, when $\theta_W>\kappa(\hat{G}_{X,\beta}) \frac{1 + \beta^\prime \beta+ \left\| \hat{Q} \right\|}{\beta^\prime \beta \left\| \hat{Q} \right\|} \left\| \hat{G}_{U,\mu} \right\| $
    \item an invertible matrix $\Omega$ exists, so that a subset $S$ where all the matrices $W(U,\mu)$ arising from pairs $(U,\mu)$ are such that $\Omega W(U,\mu) \ge 0$ and $W(U,\mu) \Omega \ge 0$.
    Also, parameters $\theta_W$ and $\theta_V$ are such that $\theta_W > \kappa(\hat{G}_{X,\beta}) \kappa(\Omega) \frac{1+\beta^\prime \beta+ \left\| \hat{Q} \right\|}{\beta^\prime \beta \left\| \hat{Q} \right\|}$
\end{enumerate}

with $\sup \mu$ and $\sup F(U)$ the supremum of $\mu$ and $F(U)$ and, according to Lemma 7, 
\begin{equation}
\label{eq26}
\begin{aligned}
    \hat{G}_{X,\beta}
   \ = \;
   \begin{pmatrix}
      I & - I \bigotimes \hat{Q}  \\
      (\beta \beta^\prime)\bigotimes I  & I
    \end{pmatrix} 
    \\
    \hat{G}_{U,\mu}
    =
    I +
     \left( 1-\theta_W^2 
     \left\| W_2 \right\|_F^2 \right)^{1/2}
          \left\| I-R^\prime\left(RR^\prime\right)^{-1} R
     \right\|
    \begin{pmatrix}
        \sup F(U) \\
        \sup \mu 
    \end{pmatrix}  
    \begin{pmatrix}
        \sup F(U) \\
        \sup \mu 
    \end{pmatrix}  ^\prime
\end{aligned}
\end{equation}

Therefore, just as was found in RNNs, if certain conditions are met the parameters of constrained RNNs can be found by simple iteration. 

This result allows us to turn to the specific constraints imposed to FNNs: information flows in only one direction and there are no feedback loops. Assuming each neuron is indexed from the initial layers to the last this means the lower triangular matrix of $W$ has only zeros, including its diagonal. Therefore, in a FNN's, the optimal $W^*$ is not invertible and thus it is not evident a subset $S$ exists where convergence is assured.

The following example demonstrates that certain FNNs lack a subset for which the iteration in \ref{eq25} converges in the sense of Theorem 2. The network consists of 5 neurons arranged in three layers: neurons 1 and 2 are in the first layer, neurons 3 and 4 are in the second layer, and neuron 5 is in the third layer. Each layer is fully connected to the next, as summarized in the following matrix $W$: 
\begin{equation}
\label{eq27}
\begin{aligned}
    W^* = 
    \begin{pmatrix}
        0 & 0 & w_{1,3} & w_{1,4} & 0 \\
        0 & 0 & w_{2,3} & w_{2,4} & 0 \\
        0 & 0 & 0 & 0 & w_{3,5} \\
        0 & 0 & 0 & 0 & w_{4,5} \\
        0 & 0 & 0 & 0 & 0 
    \end{pmatrix}  
\end{aligned}
\end{equation}

Then, simple algebra shows that whenever $w_{3,5}$ and $w_{4,5}$ have opposing signs it will be impossible to find an invertible matrix $\Omega$ where $(\Omega W^* \ge 0$ and $W^* \Omega \ge 0$ (entry wise).

It is known that the structure of connections may help or hinder the estimation of parameters. For example, studies have found that adding <<skip>> or <<shortcut>> connections produce loss functions that train easier (\cite{li2018visualizing, he2016deep}). According to our model, incorporating feedback loops would further simplify training, even if the weights describing these loops were fixed (conserving the number of free parameters constant).

\section{A simple application: regressing polinomials}

In this section, we describe the implementation of a toy model where we employ Recurrent Neural Networks (RNNs) to regress a polynomial function. The primary goal is to demonstrate that Optimal Contraction can effectively capture the main features of a polynomial. The results validate the model's capability to approximate the polynomial's behavior accurately.

In particular, we regress the polinomial $y=x^3+x^2-10x$ in the domain $[-5,5]$, with 50 evenly distributed data points. The model includes only 3 neurons, in a complete RNN, and the raw data $X$ consists of a constant and the domain of the curve. 

We choose the softplus as activation function $f(c) = \frac{1}{\alpha} ln(1+e^{\alpha c})$, with $\alpha = 0.05$ constants $\theta_W = 1.2$, $\theta_V = 0.05$, $\beta = \begin{pmatrix} 1 & 1 & 1 \end{pmatrix}'$, $b = \begin{pmatrix} 0 & 1 & 2 \end{pmatrix}'$ and $\delta = 0.001$. The method ends if $20.000$ iterations are repeated or the squared sum of difference between estimated parameters in each iteration is less than $0.001$.

As a result, we are able find matrices $W$ and $V$. In particular, we find:\\
$W = 
    \begin{pmatrix}
         0.22232453  &  0.14096028 & -0.10443518 \\
        -0.06913344  & 0.1243412  & -0.35466878 \\
         0.12218206 & -0.11587928  & 0.66321421
    \end{pmatrix}
$ \\
$V = 
    \begin{pmatrix}
        -78.97621134 & -105.97137723  & 62.51969179 \\
        80.25486682  & -44.77352714  & -15.69360293
    \end{pmatrix}
$

As can be observed in the following figures, the method is able to closely approximate the polinomial, and the algorithm rapidly converges towards the optimal solution:

\begin{figure}[H]
\centering
  \begin{minipage}[b]{0.4\textwidth}
    \includegraphics[width=\textwidth]{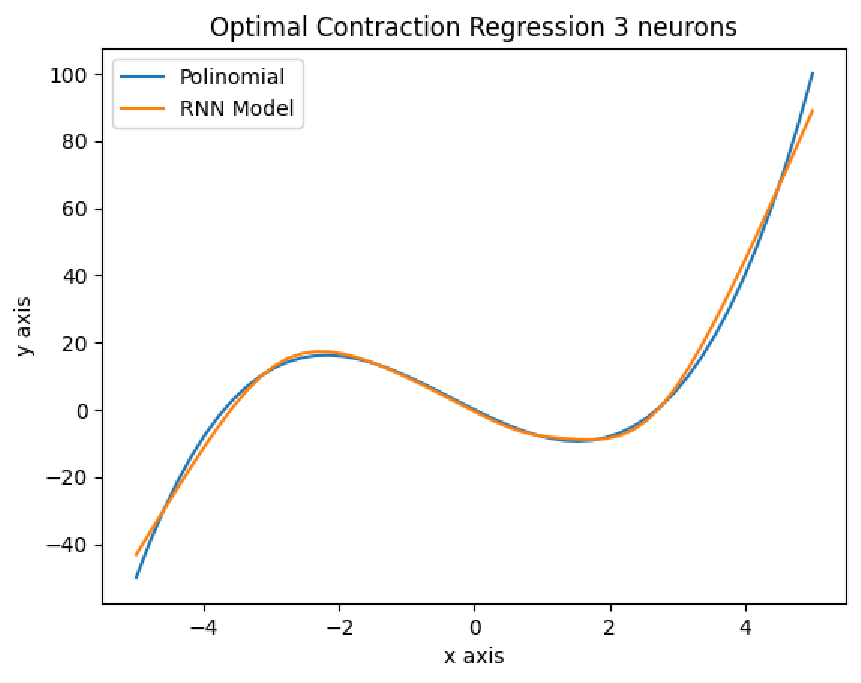}
    \caption{Model vs Data}
  \end{minipage}
  \hfill
  \begin{minipage}[b]{0.4\textwidth}
    \includegraphics[width=\textwidth]{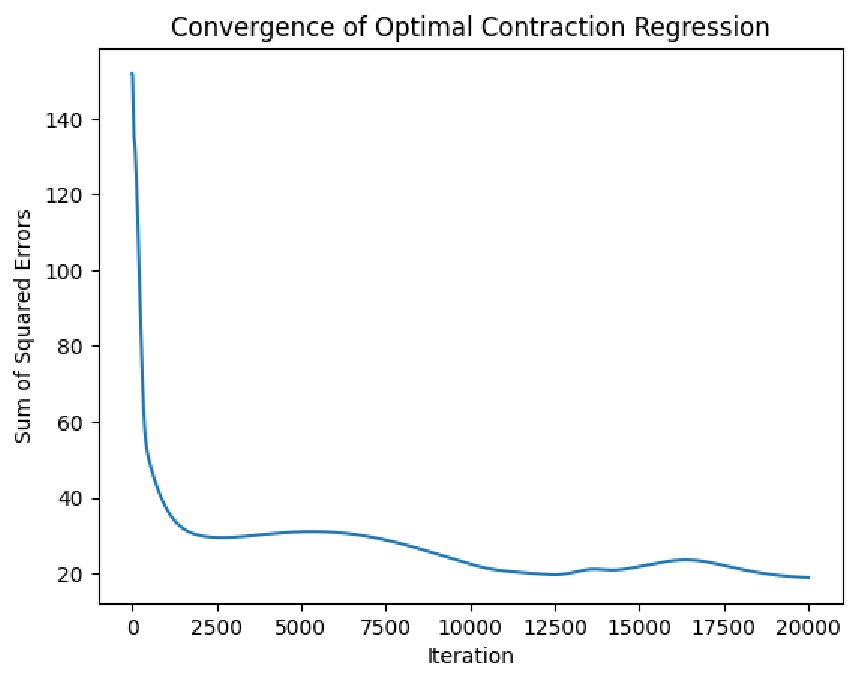}
    \caption{Sum of Squared Errors}
  \end{minipage}
\end{figure}

The results confirm that Optimal Contraction can effectively regress polynomial functions by iteratively capturing their features. This success showcases the potential of Optimal Contraction for function approximation tasks.

\newpage
\section{Conclusions}
This paper introduces an alternative approach for estimating the parameters of Recurrent and Feedforward Neural Networks. From a theoretical standpoint, we identify conditions under which parameter estimation by simple iteration converges. 

The proposed model is deeply rooted in the principles of Classical Statistics, thereby facilitating the application of its extensive toolkit. For instance, the computation of p-values for parameter estimates can be accomplished using Fisher's Information Matrix. This integration of Classical Statistical methods with Neural Network parameter estimation not only enhances the robustness of the parameters but also provides a rigorous statistical framework for their evaluation.

The methodology described in this study is applied to a prediction task. However, its utility may extend beyond regression. Further research could adapt it to estimate parameters in other machine learning paradigms, such as clustering and classification, making it a versatile approach in the field of machine learning.

Finally, Artificial Neural Networks draw inspiration from biological brains. It remains to be answered whether the human brain re calibrates the synaptic strength between neurons through a mechanism analogous to this paper's proposed algorithm. Investigating this hypothesis in future research could significantly deepen our understanding of neural plasticity and the underlying processes of learning. Such insights would not only advance the field of neuroscience but could also lead to the development of more sophisticated and biologically plausible models of artificial intelligence. 

\newpage
\printbibliography

\renewcommand\theequation{\Alph{section}\arabic{equation}} 
\counterwithin*{equation}{section} 
\renewcommand\thefigure{\Alph{section}\arabic{figure}} 
\counterwithin*{figure}{section} 
\renewcommand\thetable{\Alph{section}\arabic{table}} 
\counterwithin*{table}{section} 

\begin{appendices}

\newpage
\section{Lemmas}

\subsection{Lemma 1}

The first order condition of \ref{eq8} implies:
\begin{equation}
        W=\frac{1}{\theta_W} \frac{F\left(U\right)^\prime\mu}{({1+ \left \| F\left(U\right)^\prime\mu \right \|_F^2})^{\frac{1}{2}}}
\end{equation}

with  $\left \| F\left(U\right)^\prime\mu \right \|^2=tr\left(\mu^\prime F\left(U\right)F\left(U\right)^\prime\mu\right)$

Demonstration:
The first order condition of \ref{eq8} with respect to $W$ is:

\begin{equation}
    \theta_W\left(1-\theta_W^2tr\left(W^\prime W\right)\right)^{-\frac{1}{2}}W=F\left(U\right)^\prime\mu
\end{equation}

Therefore

\begin{equation}
    \frac{\theta_W^2Tr\left(W^\prime W\right)}{1-\theta_W^2Tr\left(W^\prime W\right)}=tr\left(\mu^\prime F\left(U\right)F\left(U\right)^\prime\mu\right)
\end{equation}

\begin{equation}
   \theta_W^2Tr\left(W^\prime W\right)=tr\left(\mu^\prime F\left(U\right)F\left(U\right)^\prime\mu\right)\left(1-\theta_W^2Tr\left(W^\prime W\right)\right)
\end{equation}

\begin{equation}
    \theta_W^2\left(1+tr\left(\mu^\prime F\left(U\right)F\left(U\right)^\prime\mu\right)\right)Tr\left(W^\prime W\right)=tr\left(\mu^\prime F\left(U\right)F\left(U\right)^\prime\mu\right)
\end{equation}

\begin{equation}
   Tr\left(W^\prime W\right)=\frac{1}{\theta_W^2}\frac{tr\left(\mu^\prime F\left(U\right)F\left(U\right)^\prime\mu\right)}{1+tr\left(\mu^\prime F\left(U\right)F\left(U\right)^\prime\mu\right)}
\end{equation}

Now we return to the first order equation of \ref{eq8}

\begin{equation}
    W=\frac{1}{\theta_W}F\left(U\right)^\prime\mu\left(1-\theta_W^2Tr\left(W^\prime W\right)\right)^\frac{1}{2}
\end{equation}

\begin{equation}
   W=\frac{1}{\theta_W}\frac{F\left(U\right)^\prime\mu}{\left(1+tr\left(\mu^\prime F\left(U\right)F\left(U\right)^\prime\mu\right)\right)^\frac{1}{2}}
\end{equation}

\begin{equation}
   W=\frac{1}{\theta_W} \frac{F\left(U\right)^\prime\mu}{({1+ \left \| F\left(U\right)^\prime\mu \right \|_F^2})^{\frac{1}{2}}}
\end{equation}
QED.

\newpage
\subsection{Lemma 2}
In this Lemma we borrow from the Linear Algebra notation, leaving aside for the moment the Regression Analysis notation. 

The solution to the following system of equations:
\begin{equation}
    \begin{aligned}
        X_1 = A X_2 + B_1 \\
        X_2 = -X_1 c c^\prime + B_2
    \end{aligned}
\end{equation}
is
\begin{equation}
    \begin{aligned}
        X_1= B_1 - (I+c^\prime c A)^{-1} A B_1 c c^\prime + A B_2 - (I+c^\prime c A)^{-1} A^2 B_2 c c^\prime  \\
        X_2= B_2 - (I+c^\prime c A)^{-1} A B_2 c c^\prime - B_1 c c^\prime + c^\prime c (I+c^\prime c A)^{-1} A B_1 c c^\prime
    \end{aligned}
\end{equation}

Demonstration: \\
The system of equations is of the form

\begin{equation}
    \begin{aligned}
        X_1 = A X_2 + B_1 \\
        X_2 = -X_1 c c^\prime + B_2
    \end{aligned}
\end{equation}

The system can be re-written as

\begin{equation}
    \begin{aligned}
        X_1 + A X_1 c c^\prime = D_1 = B_1 + A B_2  \\
        X_2 + A X_2 c c^\prime= D_2 = B_2 - B_1 c c^\prime
    \end{aligned}
\end{equation}

That is, we must solve two Sylvester equations of the type $ X + A X c c^\prime = D$

Using the Kronecker product we can re-write the equation as
\begin{equation}
    \begin{aligned}
        (I + c c^\prime \bigotimes A) vec(X) = vec (D)
    \end{aligned}
\end{equation}

So that $vec(X)=(I + c c^\prime \bigotimes A)^{-1} vec(D)$

Fortunately, that particular inverse has an analytical solution
\begin{equation}
    \begin{aligned}
        (I + c c^\prime \bigotimes A)^{-1} = I - c c^\prime \bigotimes \left( (I+c^\prime c A)^{-1} A \right)
    \end{aligned}
\end{equation}
Which means
\begin{equation}
    \begin{aligned}
        vec(X)= vec(D) - c c^\prime \bigotimes \left( (I+c^\prime c A)^{-1} A \right) vec (D)
    \end{aligned}
\end{equation}
So that, by taking $vec^{-1}()$ on both sides we find
\begin{equation}
    \begin{aligned}
        X= D - (I+c^\prime c A)^{-1} A D c c^\prime  
    \end{aligned}
\end{equation}

Therefore, our system of equations can be solved and its solution is
\begin{equation}
    \begin{aligned}
        X_1= D_1 - (I+c^\prime c A)^{-1} A D_1 c c^\prime  \\
        X_2= D_2 - (I+c^\prime c A)^{-1} A D_2 c c^\prime
    \end{aligned}
\end{equation}

Or, using the initial variables:

\begin{equation}
    \begin{aligned}
        X_1= B_1 - (I+c^\prime c A)^{-1} A B_1 c c^\prime + A B_2 - (I+c^\prime c A)^{-1} A^2 B_2 c c^\prime  \\
        X_2= B_2 - (I+c^\prime c A)^{-1} A B_2 c c^\prime - B_1 c c^\prime + c^\prime c (I+c^\prime c A)^{-1} A B_1 c c^\prime
    \end{aligned}
\end{equation}

Now lets turn to the problem of Lipschitz continuity.\\
As was observed, the original system of equations can be re-written in term of the $vec()$ operator and Kronecker product:

\begin{equation}
    \begin{aligned}
        \begin{pmatrix}
            vec(X_1) \\
            vec(X_2)
        \end{pmatrix}
        =
        \begin{pmatrix}
           (I + c c^\prime \bigotimes A)^{-1} & 0
           \\
           0 & (I + c c^\prime \bigotimes A)^{-1} 
        \end{pmatrix}
        \begin{pmatrix}
            I & I \bigotimes A
           \\
           -c c^\prime \bigotimes I & I
        \end{pmatrix}
        \begin{pmatrix}
            vec(B_1) \\
            vec(B_2)
        \end{pmatrix}
    \end{aligned}
\end{equation}

Now, we know that

\begin{equation}
    \begin{aligned}
        \begin{pmatrix}
            I & I \bigotimes A
           \\
           -c c^\prime \bigotimes I & I
        \end{pmatrix}
        =
        \begin{pmatrix}
            I & 0 \\
            0 & I
        \end{pmatrix}
        +
        \begin{pmatrix}
            0 & I \bigotimes A
           \\
           -c c^\prime \bigotimes I & 0
        \end{pmatrix}
    \end{aligned}
\end{equation}

And thus
\begin{equation}
    \begin{aligned}
    \left\|
        \begin{pmatrix}
            I & I \bigotimes A
           \\
           -c c^\prime \bigotimes I & I
        \end{pmatrix}
    \right\|
        \le
        1 +
        \left\|
        \begin{pmatrix}
            0 & I \bigotimes A
           \\
           -c c^\prime \bigotimes I & 0
        \end{pmatrix}
        \right\|
        \le
        1 + c^\prime c+ \left\| A \right\|
    \end{aligned}
\end{equation}

Also, 
\begin{equation}
    \begin{aligned}
    \left\|
        \begin{pmatrix}
           (I + c c^\prime \bigotimes A)^{-1} & 0
           \\
           0 & (I + c c^\prime \bigotimes A)^{-1} 
        \end{pmatrix}
    \right\|
        =
        \left\|
        (I + c c^\prime \bigotimes A)^{-1}
        \right\|
        = \frac{\kappa(I + c c^\prime \bigotimes A)}{\left\|
        I + c c^\prime \bigotimes A
        \right\|}
    \end{aligned}
\end{equation}

where $\kappa()$ is the condition number. Also notice that for matrices $A$ that are positive semi-definite $\left\|
        I + c c^\prime \bigotimes A
        \right\| \le \left\|
        c c^\prime \bigotimes A
        \right\| = c^\prime c \left\| A \right\|$ . 

Using the sub-multiplicative property of matrix norms we find

\begin{equation}
    \begin{aligned}
    \left\|
        \begin{pmatrix}
            vec(X_1) \\
            vec(X_2)
        \end{pmatrix}
    \right\|
        \le
    \kappa(I + c c^\prime \bigotimes A)
    \frac{1 + c^\prime c+ \left\| A \right\|}
    {c^\prime c\left\| A \right\|}
    \left\|
        \begin{pmatrix}
            vec(B_1) \\
            vec(B_2)
        \end{pmatrix}
    \right\|
    \end{aligned}
\end{equation}

which also means that:

\begin{equation}
    \begin{aligned}
    \left\|
        \begin{pmatrix}
            X_{1,1} - X_{2,2} \\
            X_{2,1} - X_{2,2}
        \end{pmatrix}
    \right\|
        \le
    \kappa(I + c c^\prime \bigotimes A)
    \frac{1 + c^\prime c+ \left\| A \right\|}
    {c^\prime c\left\| A \right\|}
    \left\|
        \begin{pmatrix}
            B_{1,1} - B_{2,2} \\
            B_{2,1} - B_{2,2}
        \end{pmatrix}
    \right\|
    \end{aligned}
\end{equation}

\newpage
\subsection{Lemma 3}

Variables $\mu$, $U$ and $W$ are bounded according to:

\begin{equation}
    \begin{aligned}
        \left\|W\right\| \leq \sup W \\
        \left\|\mu\right\|\leq \sup \mu \\
        \left\|U\right\|\leq \sup U
    \end{aligned}
\end{equation}

where

\begin{equation}
    \begin{aligned}
        \sup W = \frac{1}{\theta_W} \\
        \sup \mu = \frac{\theta_W}{\theta_W-1}\left\|Y-(b^\prime \beta) 1\right\|\left\|\beta\right\| \\
        \sup U = \left (\frac{\theta_W}{\theta_W-1}\right)^{2} \frac{1}{\theta_V}\left\|X^\prime X\right\| \left\|Y-(b^\prime \beta) 1\right\|\left\|\beta\right\|+ \frac{\theta_W}{\theta_W-1} \left\|1 b^\prime\right\|\ + \frac{1}{\theta_W-1} \left\|F(0)\right\|\
    \end{aligned}
\end{equation}

Demonstration:
First, we observe that errors are bounded.
If $V=0$ and $W=0$ then $U=1 b^\prime$ . Therefore, the Lagrangian at the minimum must be less or equal the value of the Lagrangian evaluated at that particular point in space. Moreover, the Lagrangian is the sum of three functions that are always positive and therefore any of those functions is smaller than that value. 

In particular, we find

\begin{equation}
\frac{1}{2}\varepsilon^\prime\varepsilon\le\frac{1}{2}\varepsilon^\prime\varepsilon+\frac{1}{2\theta_W}\left(1-\left(1-\theta_W^2Tr\left(W^\prime W\right)\right)^\frac{1}{2}\right)+\frac{\theta_V}{2}Tr\left(V^\prime V\right)\le\frac{1}{2} \left\|Y-(b^\prime \beta) 1\right\|^{2}
\end{equation}

Since $\varepsilon=Y-U\beta$ we find 
\begin{equation}
    \left\|Y-U \beta\right\|\le \left\|Y-(b^\prime \beta) 1\right\|
\end{equation}

Now we review bounds for each variable:

a) $W$

Calculating the Frobenius norm of $W$ from \ref{eq8} we find

\begin{equation}
    \left\|W\right\|_F^2 = \frac{1}{\theta_W^2} \frac{\left \| F\left(U\right)^\prime\mu \right \|_F^2}{1+\left \| F\left(U\right)^\prime\mu \right \|_F^2} \le \frac{1}{\theta_W^2}
\end{equation}

Now, since $\left\| W \right\| \le \left\| W \right\|_F $ we find $\left\| W \right\| \le \frac{1}{\theta_W}$

It should be noticed that the equality condition assures the range of $\left\|W\right\|$ is closed.

b) $\mu$

Calculating the norm of $\mu$ from \ref{eq8} we find

\begin{equation}
    \left\|\mu\right\|=\left\| \left(Y-U\beta\right)\beta^\prime+\dot{F}\left(U\right)\circ(\mu W^\prime) \right\|
\end{equation}

\begin{equation}
     \left\|\mu\right\|\le\left\| \left(Y-U\beta\right)\beta^\prime \right\|+ \left\| \dot{F}\left(U\right)\circ(\mu W^\prime) \right\|
\end{equation}

\begin{equation}
    \left\|\mu\right\|\le \left\| Y-U\beta \right\|\left\|\beta \right\|+ \left\| \dot{F}\left(U\right)\circ(\mu W^\prime) \right\|
\end{equation}
Since $\dot{F}\left(U\right)\in{[0,1]}$ we find

\begin{equation}
    \left\|\mu\right\|\le  \left\|Y-(b^\prime \beta) 1\right\| \left\|\beta \right\|+ \left\| \mu W^\prime \right\|
\end{equation}

\begin{equation}
    \left\|\mu\right\|\le  \left\|Y-(b^\prime \beta) 1\right\| \left\|\beta \right\|+ \frac{1}{\theta_W}\left\| \mu \right\|
\end{equation}

And therefore

\begin{equation}
    \left\|\mu\right\|\leq\frac{\theta_W}{\theta_W-1}\left\|Y-(b^\prime \beta) 1\right\|\left\|\beta\right\|
\end{equation}

c) $U$

Calculating the norm of $U$ from \ref{eq8} we find

\begin{equation}
    U=\frac{1}{\theta_V}XX^\prime\mu+1b^\prime+F\left(U\right)W
\end{equation}

\begin{equation}
    \left\|U\right\|= \left\|\frac{1}{\theta_V}XX^\prime\mu+1b^\prime+F\left(U\right)W\right\|
\end{equation}

\begin{equation}
    \left\|U\right\|\le \frac{1}{\theta_V}\left\|XX^\prime\mu\right\|+\left\|1b^\prime\right\|+\left\|F\left(U\right)W\right\|
\end{equation}

\begin{equation}
    \left\|U\right\|\le \frac{1}{\theta_V}\left\|XX^\prime\right\|\left\|\mu\right\|+\left\|1b^\prime\right\|+\frac{1}{\theta_W}\left\|F\left(U\right)-F\left(0\right)+F\left(0\right)\right\|
\end{equation}

\begin{equation}
    \left\|U\right\|\le \frac{1}{\theta_V}\left\|XX^\prime\right\|\left\|\mu\right\|+\left\|1b^\prime\right\|+\frac{1}{\theta_W}\left\|F\left(U\right)-F\left(0\right)\right\|+\frac{1}{\theta_W}\left\|F\left(0\right)\right\|
\end{equation}

And since $F()$ is Lipschitz continuous

\begin{equation}
   \left\|U\right\|\le \frac{1}{\theta_V}\left\|XX^\prime\right\|\left\|\mu\right\|+\left\|1b^\prime\right\|+\frac{1}{\theta_W}\left\|U\right\|+\frac{1}{\theta_W}\left\|F\left(0\right)\right\|
\end{equation}

\begin{equation}
       \left\|U\right\|\le \frac{\theta_W}{\theta_W-1}\frac{1}{\theta_V}\left\|XX^\prime\right\|\left\|\mu\right\|+\frac{\theta_W}{\theta_W-1}\left\|1b^\prime\right\|+\frac{1}{\theta_W-1}\left\|F\left(0\right)\right\|
\end{equation}

And substituting the result in b)

\begin{equation}
    \left\|U\right\|\leq \left (\frac{\theta_W}{\theta_W-1}\right)^{2} \frac{1}{\theta_V}\left\|X^\prime X\right\| \left\|Y-(b^\prime \beta) 1\right\|\left\|\beta\right\|+ \frac{\theta_W}{\theta_W-1} \left\|1 b^\prime\right\|\ + \frac{1}{\theta_W-1} \left\|F(0)\right\|\
\end{equation}
QED.

Also, as a corollary we find that $\sup F(U) \le \sup U + \left\|F(0)\right\|\ $

\subsection{Lemma 4}

\textbf{Lemma 4.1}

\begin{equation}
\begin{aligned}
    \left\|
   \begin{pmatrix}
         F\left(U_1\right)W(U_1,\mu_1) - F\left(U_2\right)W(U_2,\mu_2)\\
        \dot{F}\left(U_1\right)\circ(\mu_1 W^\prime(U_1,\mu_1)) - \dot{F}\left(U_2\right)\circ(\mu_2 W^\prime(U_2,\mu_2)) 
    \end{pmatrix}
    \right\|
    \le
    \frac{1}{\theta_W}
    \left\| G_{U,\mu} \right\|
     \left\|
   \begin{pmatrix}
             U_1 - U_2  \\
            \mu_1 - \mu_2 
    \end{pmatrix}  
    \right\|
    \end{aligned}
\end{equation}

with

\begin{equation}
\begin{aligned}
    G_{U,\mu}
    =
   \begin{pmatrix}
        \left( \sup F(U) \right)^2 + 1 & \sup F(U) \sup \mu \\
        \sup F(U) \sup \mu & \left( \sup \mu \right)^2 + 1
    \end{pmatrix} 
    \end{aligned}
\end{equation}

Demonstration:

First, let us notice that since $\dot{F}\left(U\right) \in [0,1]$ the difference between expressions is only damped and therefore:

\begin{equation}
\begin{aligned}
   \begin{pmatrix}
        \left\| F\left(U_1\right)W(U_1,\mu_1) - F\left(U_2\right)W(U_2,\mu_2)\right\|\\
        \left\| \dot{F}\left(U_1\right)\circ(\mu_1 W^\prime(U_1,\mu_1)) - \dot{F}\left(U_2\right)\circ(\mu_2 W^\prime(U_2,\mu_2)) \right\|
    \end{pmatrix}
    \le
    \\
   \begin{pmatrix}
        \left\| F\left(U_1\right)W(U_1,\mu_1) - F\left(U_2\right)W(U_2,\mu_2)\right\|\\
        \left\| \mu_1 W^\prime(U_1,\mu_1) - \mu_2 W^\prime(U_2,\mu_2) \right\|
    \end{pmatrix}
    \end{aligned}
\end{equation}

Now, we use that $A_1B_1 -A_2B_2 = \frac{A_1+A_2}{2}(B_1-B2) + (A_1-A_2)\frac{B_1+B_2}{2}$, then distribute and use the sub-multiplicative property of norms, finding

\begin{equation}
\begin{aligned}
   \begin{pmatrix}
        \left\| F\left(U_1\right)W(U_1,\mu_1) - F\left(U_2\right)W(U_2,\mu_2)\right\|\\
        \left\| \dot{F}\left(U_1\right)\circ(\mu_1 W^\prime(U_1,\mu_1)) - \dot{F}\left(U_2\right)\circ(\mu_2 W^\prime(U_2,\mu_2)) \right\|
    \end{pmatrix}
    \le
    \\
   \begin{pmatrix}
        \sup F\left(U\right) \left\| W(U_1,\mu_1) - W(U_2,\mu_2) \right\| + \frac{1}{\theta_W}\left\| U_1 - U_2 \right\|\\
        \sup \mu \left\| W(U_1,\mu_1) - W(U_2,\mu_2) \right\| + \frac{1}{\theta_W}\left\| \mu_1 - \mu_2 \right\|
    \end{pmatrix}  
    \end{aligned}
\end{equation}

Now, we must find bounds for $\left\| W(U_1,\mu_1) - W(U_2,\mu_2) \right\|$.\\
Let us use Lemma 1, $W(U,\mu)=\frac{1}{\theta_W}  \frac{F\left(U\right)^\prime\mu}{\left({1+ \left \| F\left(U\right)^\prime\mu \right \|^2}\right)^{\frac{1}{2}}}$ \\
Now, since $\left({1+ \left \| F\left(U\right)^\prime\mu \right \|^2}\right)^{\frac{1}{2}} \ge 1$ we find

\begin{equation}
\begin{aligned}
     \left\| W(U_1,\mu_1) - W(U_2,\mu_2) \right\| 
     \le
     \frac{1}{\theta_W}
     \left\| 
     F\left(U_1\right)^\prime\mu_1 
     -
     F\left(U_2\right)^\prime\mu_2 
     \right\|
    \end{aligned}
\end{equation}

Again, using that $A_1B_1 -A_2B_2 = \frac{A_1+A_2}{2}(B_1-B2) + (A_1-A_2)\frac{B_1+B_2}{2}$, we find
\begin{equation}
\begin{aligned}
     \left\| W(U_1,\mu_1) - W(U_2,\mu_2) \right\| 
     \le
     \frac{1}{\theta_W}
     \left(
         \sup F(U)
         \left\| 
         \mu_1 
         -
         \mu_2 
         \right\|
         +
         \sup \mu
        \left\| 
         U_1
         -
         U_2 
         \right\|
     \right)
    \end{aligned}
\end{equation}

Therefore, we find

\begin{equation}
\begin{aligned}
   \begin{pmatrix}
        \left\| F\left(U_1\right)W(U_1,\mu_1) - F\left(U_2\right)W(U_2,\mu_2)\right\|\\
        \left\| \dot{F}\left(U_1\right)\circ(\mu_1 W^\prime(U_1,\mu_1)) - \dot{F}\left(U_2\right)\circ(\mu_2 W^\prime(U_2,\mu_2)) \right\|
    \end{pmatrix}
    \le
    \\
    \frac{1}{\theta_W}
   \begin{pmatrix}
        \left( \sup F(U) \right)^2 + 1 & \sup F(U) \sup \mu \\
        \sup F(U) \sup \mu & \left( \sup \mu \right)^2 + 1
    \end{pmatrix}  
   \begin{pmatrix}
            \left\| U_1 - U_2  \right\| \\
            \left\| \mu_1 - \mu_2 \right\|
    \end{pmatrix}  
    \end{aligned}
\end{equation}

Finally, we unite both equations, finding
\begin{equation}
\begin{aligned}
    \left\|
   \begin{pmatrix}
         F\left(U_1\right)W(U_1,\mu_1) - F\left(U_2\right)W(U_2,\mu_2)\\
        \dot{F}\left(U_1\right)\circ(\mu_1 W^\prime(U_1,\mu_1)) - \dot{F}\left(U_2\right)\circ(\mu_2 W^\prime(U_2,\mu_2)) 
    \end{pmatrix}
    \right\|
    \le
    \frac{1}{\theta_W}
    \left\| G_{U,\mu} \right\|
     \left\|
   \begin{pmatrix}
             U_1 - U_2  \\
            \mu_1 - \mu_2 
    \end{pmatrix}  
    \right\|
    \end{aligned}
\end{equation}

with

\begin{equation}
\begin{aligned}
    G_{U,\mu}
    =
   \begin{pmatrix}
        \left( \sup F(U) \right)^2 + 1 & \sup F(U) \sup \mu \\
        \sup F(U) \sup \mu & \left( \sup \mu \right)^2 + 1
    \end{pmatrix} 
    \end{aligned}
\end{equation}
QED. 

\textbf{Lemma 4.2} 

Assume an optimal weight matrix $W^{*}$ exists. Also assume a matrix $\Omega$ exists, is invertible, and $\Omega W^* \ge 0$ and $W^* \Omega \ge 0$ (entry wise). 

Then, define $\Tilde{S}$ as the Subset of pairs $(U,\mu)$ where all the matrices $W(U,\mu)$ are such that $\Omega W(U,\mu) \ge 0$ and $W(U,\mu) \Omega \ge 0$ (entry wise).

Then, inside $\Tilde{S}$

\begin{equation}
\begin{aligned}
    \left\|
   \begin{pmatrix}
         F\left(U_1\right)W(U_1,\mu_1) - F\left(U_2\right)W(U_2,\mu_2)\\
        \dot{F}\left(U_1\right)\circ(\mu_1 W^\prime(U_1,\mu_1)) - \dot{F}\left(U_2\right)\circ(\mu_2 W^\prime(U_2,\mu_2)) 
    \end{pmatrix}
    \right\|
    \le
    \frac{\kappa(\Omega)}{\theta_W}
     \left\|
   \begin{pmatrix}
             U_1 - U_2  \\
            \mu_1 - \mu_2 
    \end{pmatrix}  
    \right\|
    \end{aligned}
\end{equation}

with $\kappa()$ the condition number.

Demonstration:
First, we notice $W^* \in S$, so $S$ is not void.
Then, let us notice that since $\dot{F}\left(U\right) \in [0,1]$ the difference between expressions is only damped. Second, we can re-write inequalities as:

\begin{equation}
\begin{aligned}
        F\left(U_1\right)W(U_1,\mu_1)-F\left(U_2\right)W(U_2,\mu_2) = \\
        \Bigl(
        F\left(U_1\right)W(U_1,\mu_1) \Omega-F\left(U_2\right)W(U_2,\mu_2) \Omega
        \Bigr) (\Omega)^{-1}
        \\
        \mu_1 W(U_1,\mu_1)^\prime - \mu_2 W(U_2,\mu_2)^\prime = \\
        \Bigl(
        \mu_1 (\Omega W(U_1,\mu_1))^{\prime} - \mu_2(\Omega W(U_2,\mu_2))^{\prime}
        \Bigr)(\Omega)^{-T}
    \end{aligned}
\end{equation}

That is, the condition implies $W(U_1,\mu_1) \Omega$ and $W(U_2,\mu_2) \Omega$  have the same positive (entry wise) sign. \\
We also notice that $\left\| W(U,\mu) \: \Omega\right\| \le \frac{1}{\theta_W}\left\| \Omega \right\|$, which means that $\frac{  W(U,\mu) \: \Omega  }{\frac{1}{\theta_W}\left\| \Omega \right\|} \in [0,1]$. Therefore, the pre and post multiplication of these matrices only dampens differences:

\begin{equation}
\begin{aligned}
   \begin{pmatrix}
        \left\| F\left(U_1\right)W(U_1,\mu_1) - F\left(U_2\right)W(U_2,\mu_2)\right\|\\
        \left\| \dot{F}\left(U_1\right)\circ(\mu_1 W^\prime(U_1,\mu_1)) - \dot{F}\left(U_2\right)\circ(\mu_2 W^\prime(U_2,\mu_2)) \right\|
    \end{pmatrix}
    \le
    \frac{
    \left\|
    (\Omega)^{-1}
    \right\|
    \left\|
    \Omega
    \right\|
    }{\theta_W}
   \begin{pmatrix}
            \left\| U_1 - U_2  \right\| \\
            \left\| \mu_1 - \mu_2 \right\|
    \end{pmatrix}  
    \end{aligned}
\end{equation}

So that

\begin{equation}
\begin{aligned}
    \left\|
   \begin{pmatrix}
         F\left(U_1\right)W(U_1,\mu_1) - F\left(U_2\right)W(U_2,\mu_2)\\
        \dot{F}\left(U_1\right)\circ(\mu_1 W^\prime(U_1,\mu_1)) - \dot{F}\left(U_2\right)\circ(\mu_2 W^\prime(U_2,\mu_2)) 
    \end{pmatrix}
    \right\|
    \le
    \frac{\kappa\bigl( \Omega \bigr)}{\theta_W}
     \left\|
   \begin{pmatrix}
             U_1 - U_2  \\
            \mu_1 - \mu_2 
    \end{pmatrix}  
    \right\|
    \end{aligned}
\end{equation}

with $\kappa()$ the condition number. QED.

\newpage
\subsection{Lemma 5}

Let $f(x)=(1 - \delta) g(x) +\delta A h(x)$, where $A$ is an NxN square matrix and $f(x)$, $g(x)$ and $h(x)$ are N dimensional functions of $x$, an N dimensional vector
where \\
a) $g()$ is Lipschitz continuous with a Lipschitz constant $k_g \le 1$. \\
That is, $\left\|g(x_1)-g(x_2)\right\| \le k_g \left\|x_1-x_2\right\|$, and \\
b) $h()$ is Lipschitz continuous with any Lipschitz constant $k_h$. \\
That is, $\left\|h(x_1)-h(x_2)\right\|\leq k_h \left\|x_1-x_2\right\|$ 

Then it is possible to find $\delta$ so that a $k_f(\delta)<1$ and $\left\|f(x_1)-f(x_2)\right\|\leq k_h \left\|x_1-x_2\right\|$

Demonstration
\begin{equation}
    \left\|f(x_1)-f(x_2)\right\|=\left\| (1- \delta) (g(x_1) - g(x_2)) +\delta A (h(x_1)-h(x_2)) \right\|
\end{equation}

\begin{equation}
    \left\|f(x_1)-f(x_2)\right\| 
    \le
    \left| 1- \delta \right| 
    \left\| g(x_1) - g(x_2) \right\|
    +
    \delta
    \left\|  A \right\|
    \left\|  h(x_1)-h(x_2) \right\|
\end{equation}

\begin{equation}
    \left\|f(x_1)-f(x_2)\right\| \le
      \bigl(
        k_g 
        -
        \delta 
    \left(
    k_g -
    \left\|  A \right\|
        k_h
    \right)
    \bigr)
    \left\| x_1 - x_2 \right\|
\end{equation}

Therefore, as long as $k_g > k_h \left\|A\right\|$ we can find $\delta$ so that \\ $k_f<1$, with $k_f= k_g- \delta \left( k_g - k_h \left\|  A \right\| \right) $ 

That is, when $ g(x)=x$ a contraction can be used to solve the equation given by $x = A h(x) $ for any $h(x)$ as long as $ k_h \left\|  A \right\| <1$

\subsection{Lemma 6}

\begin{equation}
    \begin{aligned}
        W(U,\mu)=\frac{1}{\theta_W} \left( \frac{1-\theta_W^2 \left\| W_2 \right\|_F^2 }{1+ \left\| W_1 \right\|_F^2}\right) W_1(U,\mu)+W_2 \\
    \end{aligned}
\end{equation}
with
\begin{equation}
    \begin{aligned}
        W_1(U,\mu)=vec^{-1}\left(\left[I-R^\prime\left(RR^\prime\right)^{-1} R\right]vec\left(F\left(U\right)^\prime\mu\right)\right)\\
        W_2=\ vec^{-1}\left(R^\prime\left(RR^\prime\right)^{-1}r\right)
    \end{aligned}
\end{equation}

Demonstration:

From the first order conditions we find:

\begin{equation}
    \begin{aligned}
        \theta_W\left(1-\theta_W^2tr\left(W^\prime W\right)\right)^{-\frac{1}{2}}W-F\left(U\right)^\prime\mu - vec^{-1}\left(R^\prime\gamma\right)=0\\
        R \: vec(W) - r = 0
    \end{aligned}
\end{equation}

To simplify equations let us first define $k(W) = \theta_W\left(1-\theta_W^2 tr\left(W^\prime W\right)\right)^{-\frac{1}{2}}$. Then

\begin{equation}
    \begin{aligned}
        W=\frac{1}{k(W)}
        \left( 
        F\left(U\right)^\prime\mu + vec^{-1}\left(R^\prime\gamma\right)
        \right)
    \end{aligned}
\end{equation}

Therefore 
\begin{equation}
    \begin{aligned}
        r=\frac{1}{k(W)} 
        R \:
        vec
        \left( 
        F\left(U\right)^\prime\mu + vec^{-1}\left(R^\prime\gamma\right)
        \right) \\
        r=\frac{1}{k(W)} 
        \left( 
        R \:
        vec \left( F\left(U\right)^\prime\mu \right) 
        + 
        R R^\prime\gamma
        \right)
    \end{aligned}
\end{equation}

Which allows us to find $\gamma$

\begin{equation}
    \begin{aligned}
        \gamma=
        (R R^\prime)^{-1}
        \left( 
        k(W) r - 
        R \:
        vec \left( F\left(U\right)^\prime\mu \right) 
        \right)
    \end{aligned}
\end{equation}

Finally, we find

\begin{equation}
    \begin{aligned}
        W=\frac{1}{k(W)}
        vec^{-1}
        \left( 
        \left( I -R^\prime (R R^\prime)^{-1} R \right)
        vec \left( F\left(U\right)^\prime\mu \right)
        \right)
        +
         vec^{-1}
        \left( R^\prime (R R^\prime)^{-1} r \right)
    \end{aligned}
\end{equation}

This proves $W=\frac{1}{k(W)}W_1 + W_2$

Now, 
\begin{equation}
    \begin{aligned}
        tr(W W^\prime) = \frac{1}{k(W)^2} tr(W_1 W_1^\prime) + tr(W_2 W_2^\prime) +\frac{2}{k(W)} tr(W_2 W_1^\prime)
    \end{aligned}
\end{equation}

But $tr(W_1 W_2^\prime)=0$ because
\begin{equation}
    \begin{aligned}
        tr(W_2 W_1^\prime) = vec (W_2)^\prime vec (W_1) \\
        = \left(R^\prime\left(RR^\prime\right)^{-1}r\right)^\prime 
        \left(\left[I-R^\prime\left(RR^\prime\right)^{-1} R \right]vec\left(F\left(U\right)^\prime\mu\right)\right) = 0
    \end{aligned}
\end{equation}

Therefore, using the definition $k(W) = \theta_W\left(1-\theta_W^2 tr\left(W^\prime W\right)\right)^{-\frac{1}{2}}$

\begin{equation}
    \begin{aligned}
        tr(W W^\prime) = \frac{1-\theta_W^2 tr\left(W^\prime W\right)}{\theta_W^2} tr(W_1 W_1^\prime) + tr(W_2 W_2^\prime)
    \end{aligned}
\end{equation}

Which allows us to find

\begin{equation}
    \begin{aligned}
        tr(W W^\prime) = \frac{1}{\theta_W^2} \frac{tr(W_1 W_1^\prime)+\theta_W^2 tr(W_2 W_2^\prime)}{1+ tr(W_1 W_1^\prime)} \\
        tr(W W^\prime) = \frac{1}{\theta_W^2} \frac{\left\|W_1 \right\|_F^2+\theta_W^2 \left\|W_2 \right\|_F^2}{1+ \left\|W_1 \right\|_F^2}
    \end{aligned}
\end{equation}

Finally, we find

\begin{equation}
    \begin{aligned}
        W=\frac{1}{\theta_W} 
        \left(
        \frac{
         1-\theta_W^2 \left\|W_2 \right\|_F^2
        }
        {
        1+ \left\|W_1 \right\|_F^2
        }
        \right)^{\frac{1}{2}}
        W_1 + W_2
    \end{aligned}
\end{equation}

It can be easily observed that if $\left\|W_2 \right\|_F < \frac{1}{\theta_W}$ then $\left\|W \right\|_F < \frac{1}{\theta_W}$. And, since $\left\|W \right\| < \left\|W \right\|_F$ this proves $\left\|W \right\| <\frac{1}{\theta_W}$ when $\left\|W_2 \right\|_F < \frac{1}{\theta_W}$.

\newpage
\subsection{Lemma 7}

\begin{equation}
\begin{aligned}
    \left\|
   \begin{pmatrix}
         F\left(U_1\right)W(U_1,\mu_1) - F\left(U_2\right)W(U_2,\mu_2)\\
        \dot{F}\left(U_1\right)\circ(\mu_1 W^\prime(U_1,\mu_1)) - \dot{F}\left(U_2\right)\circ(\mu_2 W^\prime(U_2,\mu_2)) 
    \end{pmatrix}
    \right\|
    \le
    \frac{1}{\theta_W}
    \left\| \hat{G}_{U,\mu} \right\|
     \left\|
   \begin{pmatrix}
             U_1 - U_2  \\
            \mu_1 - \mu_2 
    \end{pmatrix}  
    \right\|
    \end{aligned}
\end{equation}

with

\begin{equation}
\begin{aligned}
    \hat{G}_{U,\mu}
    =
     I
     +
     \left( 1-\theta_W^2 
     \left\| W_2 \right\|^2 \right)^{1/2}
          \left\| I-R^\prime\left(RR^\prime\right)^{-1} R
     \right\|
    \begin{pmatrix}
        \sup F(U) \\
        \sup \mu 
    \end{pmatrix}  
    \begin{pmatrix}
        \sup F(U) \\
        \sup \mu 
    \end{pmatrix}  ^\prime
    \end{aligned}
\end{equation}

Demonstration:
As in Lemma 4, let us notice that since $\dot{F}\left(U\right) \in [0,1]$ the difference between expressions is only damped and therefore:

\begin{equation}
\begin{aligned}
   \begin{pmatrix}
        \left\| F\left(U_1\right)W(U_1,\mu_1) - F\left(U_2\right)W(U_2,\mu_2)\right\|\\
        \left\| \dot{F}\left(U_1\right)\circ(\mu_1 W^\prime(U_1,\mu_1)) - \dot{F}\left(U_2\right)\circ(\mu_2 W^\prime(U_2,\mu_2)) \right\|
    \end{pmatrix}
    \le
    \\
   \begin{pmatrix}
        \left\| F\left(U_1\right)W(U_1,\mu_1) - F\left(U_2\right)W(U_2,\mu_2)\right\|\\
        \left\| \mu_1 W^\prime(U_1,\mu_1) - \mu_2 W^\prime(U_2,\mu_2) \right\|
    \end{pmatrix}
    \end{aligned}
\end{equation}

Now, we use that $A_1B_1 -A_2B_2 = \frac{A_1+A_2}{2}(B_1-B2) + (A_1-A_2)\frac{B_1+B_2}{2}$, then distribute and use the sub-multiplicative property of norms, finding

\begin{equation}
\begin{aligned}
   \begin{pmatrix}
        \left\| F\left(U_1\right)W(U_1,\mu_1) - F\left(U_2\right)W(U_2,\mu_2)\right\|\\
        \left\| \dot{F}\left(U_1\right)\circ(\mu_1 W^\prime(U_1,\mu_1)) - \dot{F}\left(U_2\right)\circ(\mu_2 W^\prime(U_2,\mu_2)) \right\|
    \end{pmatrix}
    \le
    \\
   \begin{pmatrix}
        \sup F\left(U\right) \left\| W(U_1,\mu_1) - W(U_2,\mu_2) \right\| + \frac{1}{\theta_W}\left\| U_1 - U_2 \right\|\\
        \sup \mu \left\| W(U_1,\mu_1) - W(U_2,\mu_2) \right\| + \frac{1}{\theta_W}\left\| \mu_1 - \mu_2 \right\|
    \end{pmatrix}  
    \end{aligned}
\end{equation}

where we assumed $\left\| W_2 \right\| < \frac{1}{\theta_W}$, so that also $\left\| W \right\| < \frac{1}{\theta_W}$. \\
Now, we must find bounds for $\left\| W(U_1,\mu_1) - W(U_2,\mu_2) \right\|$.

Now let us use Lemma 6, $W(U,\mu)=\frac{1}{\theta_W} \left( \frac{1-\theta_W^2 \left\| W_2 \right\|_F^2 }{1+ \left\| W_1 \right\|_F^2}\right) W_1(U,\mu)+W_2$ 

Now, since $1+ \left\| W_1 \right\|^2 \ge 1$ we find

\begin{equation}
    \begin{aligned}
     \left\| 
     W(U_1,\mu_1) - W(U_2,\mu_2) \right\| 
     \le \\
     \frac{ \left( 1-\theta_W^2 
     \left\| W_2 \right\|_F^2 \right
     )^{1/2}}{\theta_W}
     \left\| vec^{-1}
     \left(\left[I-R^\prime\left(RR^\prime\right)^{-1} R
     \right] vec \left(     
     F\left(U_1\right)^\prime\mu_1 
     -
     F\left(U_2\right)^\prime\mu_2 
     \right)
     \right)
     \right\|
    \end{aligned}
\end{equation}

which means
\begin{equation}
    \begin{aligned}
     \left\| 
     W(U_1,\mu_1) - W(U_2,\mu_2) \right\| 
     \le \\
     \frac{ \left( 1-\theta_W^2 
     \left\| W_2 \right\|_F^2 \right
     )^{1/2}}{\theta_W}
     \left\| I-R^\prime\left(RR^\prime\right)^{-1} R
     \right\|
     \left\|
     F\left(U_1\right)^\prime\mu_1 
     -
     F\left(U_2\right)^\prime\mu_2 
     \right\|
    \end{aligned}
\end{equation}

Again, using that $A_1B_1 -A_2B_2 = \frac{A_1+A_2}{2}(B_1-B2) + (A_1-A_2)\frac{B_1+B_2}{2}$, we find
\begin{equation}
\begin{aligned}
     \left\| 
     F\left(U_1\right)^\prime\mu_1 
     -
     F\left(U_2\right)^\prime\mu_2
     \right\| 
     \le
         \sup F(U)
         \left\| 
         \mu_1 
         -
         \mu_2 
         \right\|
         +
         \sup \mu
        \left\| 
         U_1
         -
         U_2 
         \right\|
    \end{aligned}
\end{equation}

Therefore, we find

\begin{equation}
\begin{aligned}
   \begin{pmatrix}
        \left\| F\left(U_1\right)W(U_1,\mu_1) - F\left(U_2\right)W(U_2,\mu_2)\right\|\\
        \left\| \dot{F}\left(U_1\right)\circ(\mu_1 W^\prime(U_1,\mu_1)) - \dot{F}\left(U_2\right)\circ(\mu_2 W^\prime(U_2,\mu_2)) \right\|
    \end{pmatrix}
    \le
    \\
     \frac{1}{\theta_W}
     \left(
     I +
     \left( 1-\theta_W^2 
     \left\| W_2 \right\|_F^2 \right)^{1/2}
          \left\| I-R^\prime\left(RR^\prime\right)^{-1} R
     \right\|
   \begin{pmatrix}
        \sup F(U) \\
        \sup \mu 
    \end{pmatrix}  
    \begin{pmatrix}
        \sup F(U) \\
        \sup \mu 
    \end{pmatrix}  ^\prime
    \right)
   \begin{pmatrix}
            \left\| U_1 - U_2  \right\| \\
            \left\| \mu_1 - \mu_2 \right\|
    \end{pmatrix}  
    \end{aligned}
\end{equation}

Finally, we find
\begin{equation}
\begin{aligned}
    \left\|
   \begin{pmatrix}
         F\left(U_1\right)W(U_1,\mu_1) - F\left(U_2\right)W(U_2,\mu_2)\\
        \dot{F}\left(U_1\right)\circ(\mu_1 W^\prime(U_1,\mu_1)) - \dot{F}\left(U_2\right)\circ(\mu_2 W^\prime(U_2,\mu_2)) 
    \end{pmatrix}
    \right\|
    \le
    \frac{1}{\theta_W}
    \left\| \hat{G}_{U,\mu} \right\|
     \left\|
   \begin{pmatrix}
             U_1 - U_2  \\
            \mu_1 - \mu_2 
    \end{pmatrix}  
    \right\|
    \end{aligned}
\end{equation}

with

\begin{equation}
\begin{aligned}
    \hat{G}_{U,\mu}
    =
     I
     +
     \left( 1-\theta_W^2 
     \left\| W_2 \right\|_F^2 \right)^{1/2}
          \left\| I-R^\prime\left(RR^\prime\right)^{-1} R
     \right\|
    \begin{pmatrix}
        \sup F(U) \\
        \sup \mu 
    \end{pmatrix}  
    \begin{pmatrix}
        \sup F(U) \\
        \sup \mu 
    \end{pmatrix}  ^\prime
    \end{aligned}
\end{equation}
QED.

\end{appendices}

\end{document}